\documentclass[journal ]{new-aiaa}
\usepackage{mwe}

\usepackage{import}
\usepackage{multirow}
\usepackage{enumerate} % To enumerate list from 0 
\usepackage{acro} % For acronym management
\usepackage{comment} % To comment
\usepackage[utf8]{inputenc} % new page
\usepackage{amsmath,bm}
\usepackage{placeins}

\DeclareAcronym{fov}{
  short = FOV ,
  long  = {Field Of View} ,
  tag = abbrev
}
\DeclareAcronym{ml}{
  short = ML ,
  long  = {Machine Learning} ,
  tag = abbrev
}
\DeclareAcronym{ai}{
  short = AI ,
  long  = {Artificial Intelligence} ,
  tag = abbrev
}
\DeclareAcronym{nn}{
  short = NN ,
  long  = {Neural Network} ,
  tag = abbrev
}
\DeclareAcronym{cnn}{
  short = CNN ,
  long  = {Convolutional Neural Network} ,
  tag = abbrev
}
\DeclareAcronym{elm}{
  short = ELM ,
  long  = {Extreme Learning Machine} ,
  tag = abbrev
}
\DeclareAcronym{celm}{
  short = CELM ,
  long  = {Convolutional Extreme Learning Machine} ,
  tag = abbrev
}
\DeclareAcronym{hcelm}{
  short = HCELM ,
  long  = {Hybrid Convolutional Extreme Learning Machine} ,
  tag = abbrev
}
\DeclareAcronym{hcelm3}{
  short = $\text{HCELM}_3$ ,
  long  = {Hybrid Convolutional Extreme Learning Machine 3} ,
  tag = abbrev
}
\DeclareAcronym{tf}{
  short = TF ,
  long  = {TensorFlow} ,
  tag = abbrev
}
\DeclareAcronym{leakyrelu}{
  short = LeakyReLU ,
  long  = {Leaky Rectified Linear Unit} ,
  tag = abbrev
}
\DeclareAcronym{relu}{
  short = ReLU ,
  long  = {Rectified Linear Unit} ,
  tag = abbrev
}
\DeclareAcronym{nrelu}{
  short = nReLU ,
  long  = {Normalized Rectified Linear Unit} ,
  tag = abbrev
}
\DeclareAcronym{tanh}{
  short = tanh ,
  long  = {hyperbolic tangent} ,
  tag = abbrev
}
\DeclareAcronym{mbgd}{
  short = MBGD ,
  long  = {Mini-Batch Gradient Descent} ,
  tag = abbrev
}
\DeclareAcronym{gd}{
  short = GD ,
  long  = {Gradient Descent} ,
  tag = abbrev
}
\DeclareAcronym{sgd}{
  short = SGD ,
  long  = {Stochastic Gradient Descent} ,
  tag = abbrev
}
\DeclareAcronym{ls}{
  short = LS ,
  long  = {Least Square} ,
  tag = abbrev
}

\DeclareAcronym{d}{
  short = D ,
  long  = {65803 Didymos} ,
  tag = abbrev
}
\DeclareAcronym{h}{
  short = H ,
  long  = {103P/Hartley} ,
  tag = abbrev
}
\DeclareAcronym{l}{
  short = L ,
  long  = {21 Lutetia} ,
  tag = abbrev
}
\DeclareAcronym{p}{
  short = P ,
  long  = {67P/Churyumov–Gerasimenko} ,
  tag = abbrev
}
\DeclareAcronym{com}{
  short = CoM ,
  long  = {Center of Mass} ,
  tag = abbrev
}
\DeclareAcronym{cof}{
  short = CoF ,
  long  = {Center of Figure} ,
  tag = abbrev
}
\DeclareAcronym{cob}{
  short = CoB ,
  long  = {Center of Brightness} ,
  tag = abbrev
}
\DeclareAcronym{ip}{
  short = IP ,
  long  = {Image Processing} ,
  tag = abbrev
}
\DeclareAcronym{imp}{
  short = ImP ,
  long  = {Image Plane} ,
  tag = abbrev
}
\DeclareAcronym{los}{
  short = LoS ,
  long  = {Line of Sight} ,
  tag = abbrev
}
\usepackage[utf8]{inputenc}
\usepackage{textcomp}
\usepackage{graphicx}
\usepackage{amsmath}
\usepackage[version=4]{mhchem}
\usepackage{siunitx}
\usepackage{ifxetex}
\usepackage{tocloft}
\usepackage{textpos}
\usepackage{tabularx,longtable,multirow,subfig,caption}
\usepackage{fancyhdr}
\usepackage{color}
\usepackage{url}
\usepackage{float}
\usepackage[english]{babel}
\usepackage{graphicx}
\usepackage{dsfont}
\usepackage{array}
\usepackage{latexsym}
\usepackage{etoolbox}
\usepackage{enumerate}
\usepackage{enumitem} 
\usepackage{titlesec}
\usepackage{afterpage}
\usepackage{algorithm}
\usepackage{algpseudocode}
\usepackage{multicol}
\usepackage{setspace}
\usepackage{eqparbox}
\usepackage{wrapfig}
\usepackage{booktabs}
\usepackage{bm}
\usepackage{soul,marginnote}
\usepackage[outdir=./]{epstopdf}
\usepackage{threeparttable}
\usepackage[table]{xcolor}

\usepackage[sort&compress,numbers]{natbib}

\usepackage{doi}

\makeatletter
\newcounter{elimination@steps}
\newcolumntype{R}[1]{>{\raggedleft\arraybackslash$}p{#1}<{$}}
\def\elimination@num@rights{}
\def\elimination@num@variables{}
\def\elimination@col@width{}

\newcommand{\eliminationstep}[2]
{
    \ifnum\value{elimination@steps}>0\leadsto\quad\fi
    \left[
        \ifnum\elimination@num@rights>0
            \begin{array}
            {@{}*{\elimination@num@variables}{R{\elimination@col@width}}
            |@{}*{\elimination@num@rights}{R{\elimination@col@width}}}
        \else
            \begin{array}
            {@{}*{\elimination@num@variables}{R{\elimination@col@width}}}
        \fi
            #1
        \end{array}
    \right]
    & 
    \begin{array}{l}
        #2
    \end{array}
    &%                                    moved second & here
    \addtocounter{elimination@steps}{1}
}
\makeatother

\makeatletter  
\def\colvec#1{\expandafter\colvec@i#1,,,,,,,,,\@nil}
\def\colvec@i#1,#2,#3,#4,#5,#6,#7,#8,#9\@nil{% 
  \ifx$#2$ \begin{bmatrix}#1\end{bmatrix} \else
    \ifx$#3$ \begin{bmatrix}#1\\#2\end{bmatrix} \else
      \ifx$#4$ \begin{bmatrix}#1\\#2\\#3\end{bmatrix}\else
        \ifx$#5$ \begin{bmatrix}#1\\#2\\#3\\#4\end{bmatrix}\else
          \ifx$#6$ \begin{bmatrix}#1\\#2\\#3\\#4\\#5\end{bmatrix}\else
            \ifx$#7$ \begin{bmatrix}#1\\#2\\#3\\#4\\#5\\#6\end{bmatrix}\else
              \ifx$#8$ \begin{bmatrix}#1\\#2\\#3\\#4\\#5\\#6\\#7\end{bmatrix}\else
                 \PackageError{Column Vector}{The vector you tried to write is too big, use bmatrix instead}{Try using the bmatrix environment}
              \fi
            \fi
          \fi
        \fi
      \fi
    \fi
  \fi 
}  
\makeatother

\robustify{\colvec}

\definecolor{LightGreyTable}{gray}{0.95}
\definecolor{MediumGreyTable}{gray}{0.75}
\definecolor{GreyTable}{gray}{0.5}

\title{Design of Convolutional Extreme Learning Machines for Vision-Based Navigation Around Small Bodies}

\author{Mattia Pugliatti \footnote{Ph.D. Student, Department of Aerospace Science and Technology, Via La Masa 34, mattia.pugliatti@polimi.it}} 

\author{Francesco Topputo\footnote{Full Professor, Department of Aerospace Science and Technology, Via La Masa 34, francesco.topputo@polimi.it, AIAA senior member}}

\affil{Politecnico di Milano, 20156, Milan, Italy}

\begin{document}

\maketitle

\begin{abstract}
Deep learning architectures such as convolutional neural networks are the standard in computer vision for image processing tasks. Their accuracy however often comes at the cost of long and computationally expensive training, the need for large annotated datasets, and extensive hyper-parameter searches. On the other hand, a different method known as convolutional extreme learning machine has shown the potential to perform equally with a dramatic decrease in training time. Space imagery, especially about small bodies, could be well suited for this method. In this work, convolutional extreme learning machine architectures are designed and tested against their deep-learning counterparts. Because of the relatively fast training time of the former, convolutional extreme learning machine architectures enable efficient exploration of the architecture design space, which would have been impractical with the latter, introducing a methodology for an efficient design of a neural network architecture for computer vision tasks. Also, the coupling between the image processing method and labeling strategy is investigated and demonstrated to play a major role when considering vision-based navigation around small bodies. 
\end{abstract}

\section{Introduction}\label{sec:introduction}
Missions towards small bodies, such as asteroids and comets, are becoming increasingly interesting for national space agencies, companies, and smaller players such as research centers and universities \cite{GNC_survey_JPL}. From a scientific point of view, these bodies gather valuable information on the Solar System’s primordial state. Their heterogeneous distribution makes them abundantly available for resource exploitation and easily accessible from Earth, which in turn might pose an existential threat to human activities and an opportunity for technology demonstration.

The capability to autonomously navigate around a known celestial body is of paramount importance to enable any autonomous decision-making process on-board a spacecraft \cite{GNC_survey_JPL}. When considering both the proximity environment of a small body and the navigation sensors available on the market, cameras are usually preferred as they are light, compact, and have low power demand. For these reasons, the use of passive cameras, in combination with \ac{ip} algorithms, provides compelling navigation performances with cost-effective hardware.

A promising family of \ac{ip} methods is represented by data-driven algorithms, especially those making use of deep learning. In this context, traditional deep learning architectures like \ac{cnn} and its variants have demonstrated their exceptional capability to extract high-level features from images and process their nonlinear mapping with labels, representing the state of the art in computer vision for several tasks \cite{Szeliski2022}. On the other hand, \ac{cnn}s often need a large amount of data for training, which occurs via \ac{gd} methods and requires substantial computational resources \cite{Szeliski2022}. 

An opposite approach to deep architectures and learning via \ac{gd} exists, which in some cases has been demonstrated to perform similarly or better. \ac{elm} is a theoretical formulation of a learning strategy that has been first introduced in \cite{Huang2006} and later organized more consistently first in \cite{Huang2014} and then in \cite{Huang2015_ELM}. In these works \ac{elm} theory is applied to single layers feed-forward networks whose weights and biases are initialized randomly. Training happens using a \ac{ls} method to adjust the weights of the connections between the single hidden layer and the output one. Because \ac{ls} is an order of magnitude faster than \ac{gd}, training happens extremely fast. The idea is that with enough randomized neurons in the hidden layers, a network would be capable to generate a multi-dimensional basis that can be used to map the nonlinear relationship between input and labels. \ac{elm} is demonstrated to perform similarly or better than deep architectures \cite{Huang2006,Huang2014,Huang2015_ELM}, requiring only a fraction of their training time. 

At the same time \ac{elm} concepts were being formalized and used, the pivotal work in \cite{DBLP:conf/icml/SaxeKCBSN11} stressed the unexpected performance achieved with \ac{cnn} when using random weights and biases in the convolutional kernels. The authors prove that: 1) A surprising fraction of performance in a \ac{cnn} can be contributed by the intrinsic properties of the architecture alone and not from the learning algorithm used; 2) Convolutional pooling architectures can be frequency selective and translation invariant, even when random weights are used; 3) A methodology that uses randomized \ac{cnn} to search the hyper-parameters within the architecture design space perform inherently better than traditional approaches. By sidestepping the time-consuming learning process, and only focusing on those architectures with superior hierarchical structures, an order of magnitude speedup in the training process is obtained. 

These two research lines come together in \cite{Huang2015_CELM}, which extends the \ac{elm} theory to \ac{cnn} with randomized kernels, introducing the concept of \ac{celm} for computer vision tasks. The convolutional layers of a \ac{cnn} are set with random weights and biases, up to the fully connected layer, whose connections with the output layer are treated as an \ac{elm} architecture and solved with a \ac{ls} method. Similarly to \ac{elm}, \ac{celm} achieves extremely fast training and accuracies that may be similar to those of \ac{cnn}.

Several other works using \ac{celm} architectures and training strategies are present in the literature. A thorough, systematic review of these is illustrated in \cite{Rodrigues2021}. Interestingly, no prior work has been focused on the adoption of \ac{celm} for on-board \ac{ip} of celestial bodies.

When considering images of a small body taken from a navigation camera, the scenery is relatively simple when compared to other computer vision domains. The background and foreground are clearly distinguishable, and the surface variations are only due to morphological characteristics (i.e. craters, boulders, etc.) which only vary under illumination conditions. This domain is fundamentally simpler than that of typical computer vision applications in urban environments, which need to account for a large variety of commonly used objects, where deep architectures are the state of the art \cite{Szeliski2022}. Moreover, previous findings in \cite{DataDriven} hint to the fact that the filtering capabilities \ac{cnn}s on images of a small body are critical in pushing the performance compared to traditional methods. 

Pivoting on these prior works and the existing gaps in the literature, in this paper the authors attempt to answer three fundamental questions, namely: 1) Although \ac{cnn} are superior over complex scenery, can simpler methods perform better when it comes to analyzing images of small bodies?; 2) As suggested in \cite{DBLP:conf/icml/SaxeKCBSN11}, can a methodology be developed to bootstrap the training of \ac{cnn} exploiting the capability of \ac{celm} to identify the most promising architectures?; 3) Which is the best labeling strategy and reference frame to be used when considering a visual-based navigation application around a small body? 

In this work, \ac{celm} is thus investigated as a possible alternative to \ac{cnn} for autonomous vision-based navigation systems around small bodies. This is done with an extensive analysis considering 4 different small body shapes and 5 different labeling strategies, for a total of 20 scenarios, each of which is examined with 4 different \ac{ip} methods, resulting in tens of thousands of different architectures explored. Such efficient exploration of the architecture design space is possible thanks to the extremely fast training time of \ac{celm}, which is orders of magnitude faster than that of \ac{cnn} \cite{Huang2015_CELM}. It is also demonstrated that the labeling strategy and reference frame play a crucial role during training and significantly affect the performance of the methods considered. We believe that \ac{celm} can represent, with the proper labeling strategy, a promising alternative in space imagery, especially the one related to small bodies. 

The rest of the paper is organized as follows. In Section \ref{sec:methodology} the three pillars sustaining the methodology are discussed in detail. These are the dataset generation, the preprocessing pipeline of the image-label pairs, and the description of the \ac{ip} methods developed and used in this paper. The performance of these methods is then compared extensively in Section \ref{sec:results} while conclusions, and future works are discussed in Section \ref{sec:conclusions}.

\section{Methodology}\label{sec:methodology}

\subsection{Dataset generation}\label{sec:Dataset}

In this section, the dataset generation procedure is illustrated. Four small bodies are considered: \ac{d}, \ac{h}, \ac{l} and \ac{p}. Using the same methodology illustrated in \cite{Pugliatti2021_unet}, artificial morphological features such as boulders and craters are inserted into the shape models of these bodies. 

\begin{figure}[hbt!]
    \centering
    \includegraphics[width=1\textwidth]{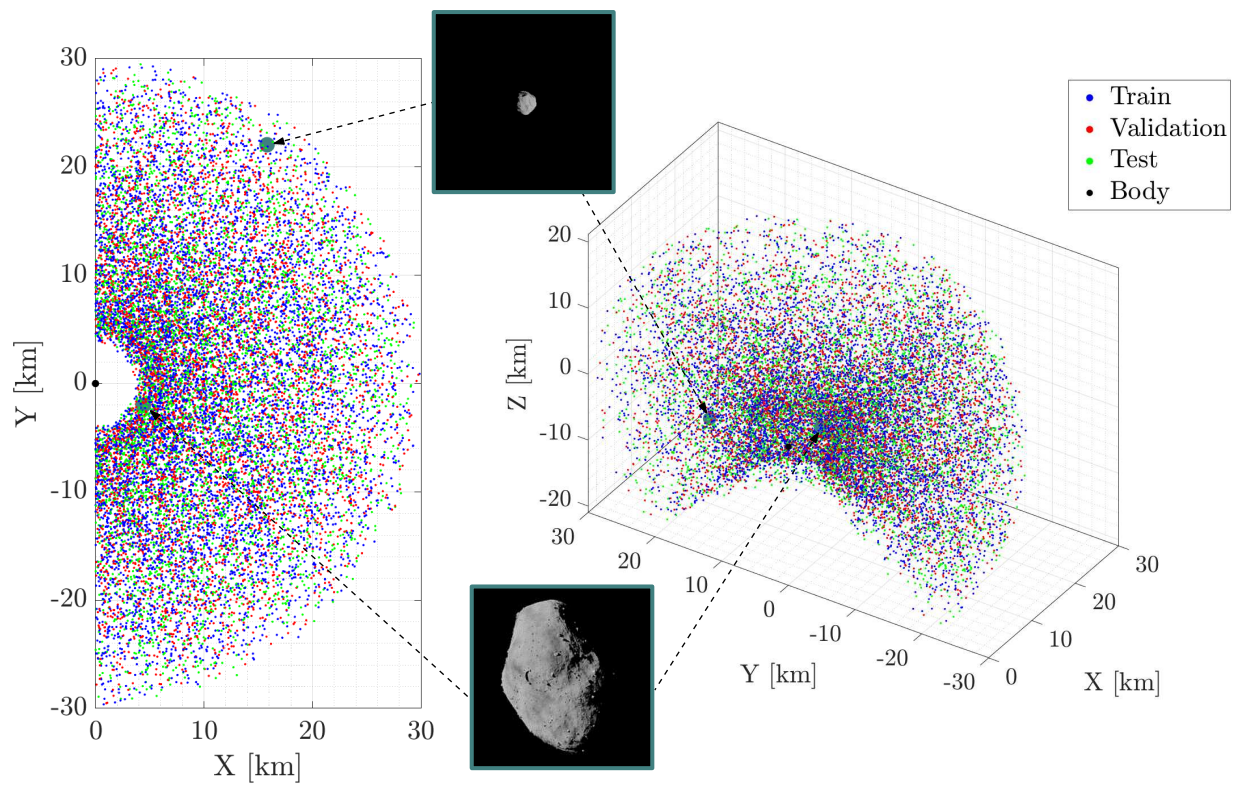}
    \caption{Cloud of $17500$ points used for training, validation, and test of the \ac{ip} methods.}
    \label{fig:datasets}
\end{figure}

For each body, a total of $17500$ grayscale images are rendered in Blender\footnote{\href{https://www.blender.org/}{https://www.blender.org/}, retrieved 25 August 2022.}. These are split into training, validation, and test sets respectively made of $7500$, $5000$, and $5000$ images. As illustrated in \autoref{fig:datasets}, these sets are randomly distributed across a cloud of points around the body in what is defined as the $W $  reference frame. This reference frame is an inertially fixed reference frame, centered on the \ac{com} of the body, with the $X$-axis oriented towards the projection of the Sun in the body equatorial plane and the $Z$-axis as the north pole of the body. Another reference frame that is used in this paper is the $AS $  one, whose axes are fixed to the body surface and are obtained from a rigid rotation around the $Z$-axis of the $W $  reference frame. The points in \autoref{fig:datasets} are distributed with range $\rho \in [5,\ 30]\ {\rm km}$, azimuth $\phi_1 \in [-90,\ 90]\ {\rm deg}$ and elevation $\phi_2 \in [-45,\ 45]\ {\rm deg}$ in the $W $  reference frame. These settings have been arbitrarily chosen as reasonable assumptions for a realistic close-proximity scenario for a mission with a passive sensor. Since each dataset is composed of the same set of points across different bodies, uniform scaling is applied to each one to make sure that different body sizes are all filling the \ac{fov} around $5$~km.

To generate the $8$-bit grayscale images, a camera with a $10\times10$ deg \ac{fov} and a sensor of $1024\times1024$ pixel is considered. All images are rendered assuming ideal pointing towards the \ac{com}. From an optical-navigation point of view, images can be represented by different sets of labels. Five of these are considered in this work to investigate the influence of this choice on the performance of the \ac{ip} methods. The first one is represented by optical observables linked with geometrical quantities directly detectable from images. In this work, these are: 
\begin{equation}
    \bm{\delta} = \bm{CoF}-\bm{CoB}
    \quad , \quad 
    \rho
\end{equation}
where $\bm{\delta}$ is the difference in pixels between the \ac{cob} and \ac{cof} of the body projected in the image plane and $\rho$ is the range from the \ac{com}. These quantities can be used to generate a position estimate in the camera frame, which can be transformed in $W $  or $AS $  by simulating on-board attitude determination from a star-tracker alongside the assumption of knowledge of the rigid rotation between the inertially fixed reference frames used by star-tracker and the $W $  or $AS $  frames. The second and third sets are represented by the spacecraft position respectively in spherical and cartesian coordinates in the $W $  frame. Similarly, the fourth and fifth sets are represented by the same coordinates in the $AS $  frame.  

By combining the 4 different small bodies with these 5 different labeling strategies, a total of 20 cases are considered in this paper. The notation used to distinguish between these cases is summarized in \autoref{tab:notation_all}. 

\begin{table}[htbp]
\caption{Notation used for the datasets used in this work.}
\label{tab:notation_all}
\centering
\begin{tabular}{l | ccc | c}
\hline
\hline
\textbf{ID} & \textbf{Body} & \textbf{Frame} & \textbf{Labels} & \textbf{Notation} \\ \hline
 1 & D & - & $\bm{\delta}, \rho$ & D1 \\
 2 & H & - & $\bm{\delta}$, $\rho$ & H1 \\
 3 & L & - & $\bm{\delta}$, $\rho$ & L1 \\
 4 & P & - & $\bm{\delta}$, $\rho$ & P1 \\
 \hline
 5 & D & $AS $  & $\phi_1,\phi_2,\rho$ & D2 \\
 6 & H & $AS $  & $\phi_1,\phi_2,\rho$ & H2 \\
 7 & L & $AS $  & $\phi_1,\phi_2,\rho$ & L2 \\
 8 & P & $AS $  & $\phi_1,\phi_2,\rho$ & P2 \\
 \hline
 9 & D & $AS $  & $X, Y, Z$ & D3 \\
 10 & H & $AS $  & $X, Y, Z$ & H3 \\ 
 11 & L & $AS $  & $X, Y, Z$ & L3 \\
 12 & P & $AS $  & $X, Y, Z$ & P3 \\
 \hline
 13 & D & $W $  & $\phi_1,\phi_2,\rho$ & D4 \\
 14 & H & $W $  & $\phi_1,\phi_2,\rho$ & H4 \\
 15 & L & $W $  & $\phi_1,\phi_2,\rho$ & L4 \\
 16 & P & $W $  & $\phi_1,\phi_2,\rho$ & P4 \\
 \hline
 17 & D & $W $  & $X, Y, Z$ & D5 \\
 18 & H & $W $  & $X, Y, Z$ & H5 \\
 19 & L & $W $  & $X, Y, Z$ & L5 \\
 20 & P & $W $  & $X, Y, Z$ & P5 \\
\hline
\hline
\end{tabular}
\end{table}

\subsection{Preprocessing}\label{sec:preprocessing}
After the dataset generation step, a total of $70000$, $1024\times1024$ grayscale images with ideal pointing towards the \ac{com} of each body are obtained. These images, however, cannot be used directly as input of the \ac{ip} methods considered in this work for three main reasons.

First, the original image resolution is too high. Due to hardware limitations, image size needs to be reduced. This is typical of the data-driven \ac{ip} methods used, which could encounter memory or processing saturation issues if working with images at native resolutions. Second, the ideal pointing assumed during rendering simplifies image generation but causes poor variability of the input-label relationship, which can cause poor generalization capability of the \ac{ip} methods. Third, rendered images are ideal, thus far from realistic camera acquisitions. 

\begin{figure}[hbt!]
    \centering
    \includegraphics[width=0.7\textwidth]{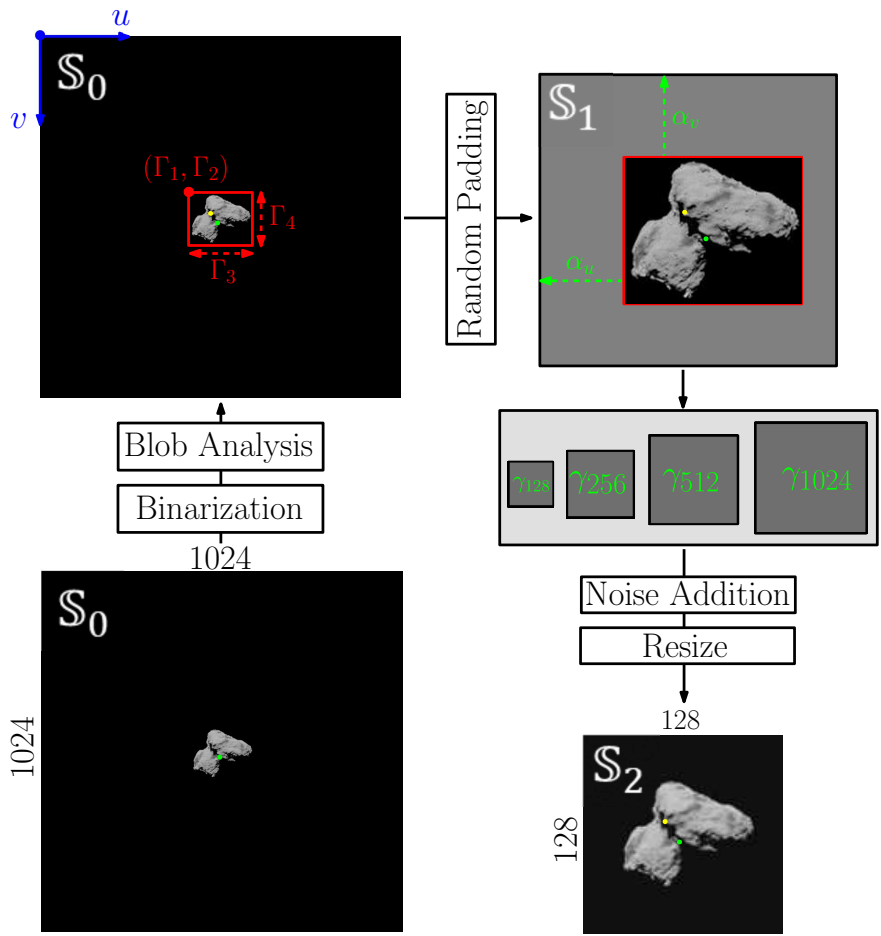}
    \caption{Sketch of the preprocessing pipeline used to transform images from $\mathbb{S}_0$ to $\mathbb{S}_2$.}
    \label{fig:IP_regularization}
\end{figure}

All three issues are addressed together in a unique preprocessing pipeline which is a novel contribution of this work. The pipeline transforms an image and its associated labels from a geometrical and rendering space referred to as $\mathbb{S}_0$ into a new space $\mathbb{S}_2$ that can be efficiently used by data-driven \ac{ip} algorithms. A sketch of the pipeline is illustrated in \autoref{fig:IP_regularization} for clarity.

The starting point of the pipeline is the image rendered in Blender, which by definition belongs to $\mathbb{S}_0$ space. The image is binarized using the Otsu method \cite{OtsuMethod} and a simple blob analysis is performed to compute the \ac{cob} (represented by the yellow point in \autoref{fig:IP_regularization}) and the bounding box $\Gamma$ (defined by two corner coordinates $\Gamma_1, \Gamma_2$, its width $\Gamma_3$, and height $\Gamma_4$). All these quantities are computed in the $UV $  frame \cite{OPNAV_tutorial}, which is considered in the image plane as centered on the top-left corner of the image. 

Cropping is performed around $\Gamma$ to transform the image to a $\Gamma_3\times\Gamma_4$ snippet. Because the \ac{ip} methods considered in this work assume fixed-size inputs, the variable-size snippet needs to be transformed into a fixed-size one. 

The upper closest possible value $\gamma$ between the largest size of $\Gamma_3\times\Gamma_4$ and $128$, $256$, $512$, or $1024$ is computed. Once the target size is determined, padding is performed using two parameters, $\alpha_u$, and $\alpha_v$, which are randomly generated. These represent the necessary integer lengths that need to be added to $\Gamma$ and are chosen respectively as $0\leq \alpha_u \leq \gamma-\Gamma_3$ and $0 \leq \alpha_v \leq \gamma-\Gamma_4$. After random padding, the image is said to be transformed in $\mathbb{S}_1$ space. Note that the padding is performed by re-using the same pixels from the image defined in $\mathbb{S}_0$ or by introducing zero-pixels values (using the nearest criteria) whenever this is not possible. When passing from $\mathbb{S}_0$ to $\mathbb{S}_1$ also the labels linked to the image need to be transformed, resulting in a change in the $\bm{CoB}$  and  $\bm{CoF}$ coordinates:  

\begin{equation}\label{eq:fromS0toS1}
    \bm{CoB}^{\mathbb{S}_1} = \bm{CoB}^{\mathbb{S}_0}-
    \begin{bmatrix}\Gamma_1\\\Gamma_2\end{bmatrix}
    +\begin{bmatrix}\alpha_u\\\alpha_v \end{bmatrix}
    \quad , \quad
    \bm{CoF}^{\mathbb{S}_1} = \bm{CoF}^{\mathbb{S}_0}-    
    \begin{bmatrix}\Gamma_1\\\Gamma_2\end{bmatrix}
    +\begin{bmatrix}\alpha_u\\\alpha_v \end{bmatrix}
\end{equation}

while the other labels remain unchanged: $\bm{\delta}^{\mathbb{S}_1} = \bm{\delta}^{\mathbb{S}_0}$; $\rho^{\mathbb{S}_1} = \rho^{\mathbb{S}_0}$; $[X,Y,Z]^{\mathbb{S}_1} = [X,Y,Z]^{\mathbb{S}_0}$; and $[\phi_1,\phi_2]^{\mathbb{S}_1} = [\phi_1,\phi_2]^{\mathbb{S}_0}$. Noise is then added to the image, which is resized to a $128\times128$ resolution, transforming it in $\mathbb{S}_2$ space. The noise step is optional for images in the datasets, while in an operational scenario it would be avoided. When passing from $\mathbb{S}_1$ to $\mathbb{S}_2$ the labels are transformed as follows: 

\begin{equation}\label{eq:fromS1toS2_p1}
    \bm{CoB}^{\mathbb{S}_2} = \bm{CoB}^{\mathbb{S}_1}\frac{128}{\gamma}
\quad , \quad
    \bm{CoF}^{\mathbb{S}_2} = \bm{CoF}^{\mathbb{S}_1}\frac{128}{\gamma}
\quad , \quad
\bm{\delta}^{\mathbb{S}_2} = \bm{\delta}^{\mathbb{S}_1}\frac{128}{\gamma}
\end{equation}

\begin{equation}\label{eq:fromS1toS2_p2}
\rho^{\mathbb{S}_2} = \rho^{\mathbb{S}_1}\frac{128}{\gamma}
\end{equation}

\begin{equation}\label{eq:fromS1toS2_p3}
[X,Y,Z]^{\mathbb{S}_2} = [X,Y,Z]^{\mathbb{S}_1}\frac{128}{\gamma}
\end{equation}

while $\phi_1$ and $\phi_2$ remain unchanged: $[\phi_1,\phi_2]^{\mathbb{S}_2} = [\phi_1,\phi_2]^{\mathbb{S}_1}$. Combining \autoref{eq:fromS0toS1} with \autoref{eq:fromS1toS2_p1}, \autoref{eq:fromS1toS2_p2}, and \autoref{eq:fromS1toS2_p3} it is possible to transform the image-label pairs for all the datasets in \autoref{tab:notation_all}. All the \ac{ip} methods used in this work use the image-label pairs in $\mathbb{S}_2$ space, which need transformation to $\mathbb{S}_0$ for on-board usage. The methods working on the datasets labeled with cartesian and polar coordinates either in the $AS $  or $W $  reference frames generate directly a position estimate as: 

\begin{equation}
    \bm{p}^{i, \mathbb{S}_0}_{est} = 
    \frac{\gamma}{128}
    \begin{bmatrix}
        X\\
        Y\\
        Z
    \end{bmatrix}^{i,\mathbb{S}_2}_{est}
\end{equation}

\begin{equation}
    \bm{p}^{i, \mathbb{S}_0}_{est} 
    = 
    \Omega\left(
    \begin{bmatrix}
        \phi_1\\
        \phi_2\\
        \rho\frac{\gamma}{128}
    \end{bmatrix}^{i,\mathbb{S}_2}_{est}
    \right)
\end{equation}

where $i$ reflects the reference frame used, and $\Omega$ is the transformation function from spherical to cartesian coordinates. On the other hand, the methods working with the ($\bm{\delta}, \rho$) labels need intermediate steps to generate a position estimate. In inference, these methods generate the following optical observables: 

\begin{equation}
    \bm{o}^{uv, \mathbb{S}_0}_{est} = 
    \begin{bmatrix}
        {CoF}^{\mathbb{S}_0}_{est,u}\\
        {CoF}^{\mathbb{S}_0}_{est,v}\\
        1
    \end{bmatrix}
    = 
    \begin{bmatrix}
        {CoB}^{\mathbb{S}_0}_u+{\delta_u}^{\mathbb{S}_2}\frac{\gamma}{128}\\
        {CoB}^{\mathbb{S}_0}_v+{\delta_v}^{\mathbb{S}_2}\frac{\gamma}{128}\\
        1
    \end{bmatrix}
    \quad , \quad
    \rho^{\mathbb{S}_0}_{est} =         \rho^{\mathbb{S}_2}\frac{\gamma}{128}
\end{equation}
where $\bm{\delta}^{\mathbb{S}_2}$ and $\rho^{\mathbb{S}_2}$ are output of the \ac{ip} methods while $\bm{CoB}^{\mathbb{S}_0}$ and $\gamma$ are parameters computed during the image preprocessing algorithm. In this work, the latter quantities are computed offline, however, it is noted that the algorithm can work also online during inference to prepare any incoming image with the proper format for the application of the \ac{ip} method. The observable vector $\bm{o}^{uv, \mathbb{S}_0}_{est}$ is transformed from the $UV $  reference frame which express pixel coordinates on the image, to the \ac{imp} reference frame using the inverse of the camera calibration matrix $\bm{K^{-1}}$ \cite{OPNAV_tutorial}: 

\begin{equation}
\bm{o}^{ImP}_{est}
=
\bm{K^{-1}}\bm{o}^{uv, \mathbb{S}_0}_{est}
\end{equation}

The $\bm{o}^{ImP}_{est}$ vector is then transformed into a \ac{los} vector in the $CAM $ reference frame. Using the attitude quaternion of the spacecraft (which is assumed to be known from attitude determination from a Star-tracker) and assuming to know the rigid rotation between the inertial reference frame used by the star-tracker and a known asteroid frame, this \ac{los} is transformed with the use of $\rho^{\mathbb{S}_0}_{est}$ into a position estimate in $W $  reference frame:

\begin{equation}
\bm{p}^{W,\mathbb{S}_0}_{est}= \bm{q}_{CAM\rightarrow W}\bm{p}^{CAM}_{est}
\end{equation}

In this work, the $W $  frame is used for simplicity and no error is simulated on the attitude quaternion.

\subsection{Image Processing}\label{sec:neural}
\begin{figure}[hbt!]
    \centering
    \includegraphics[width=1\textwidth]{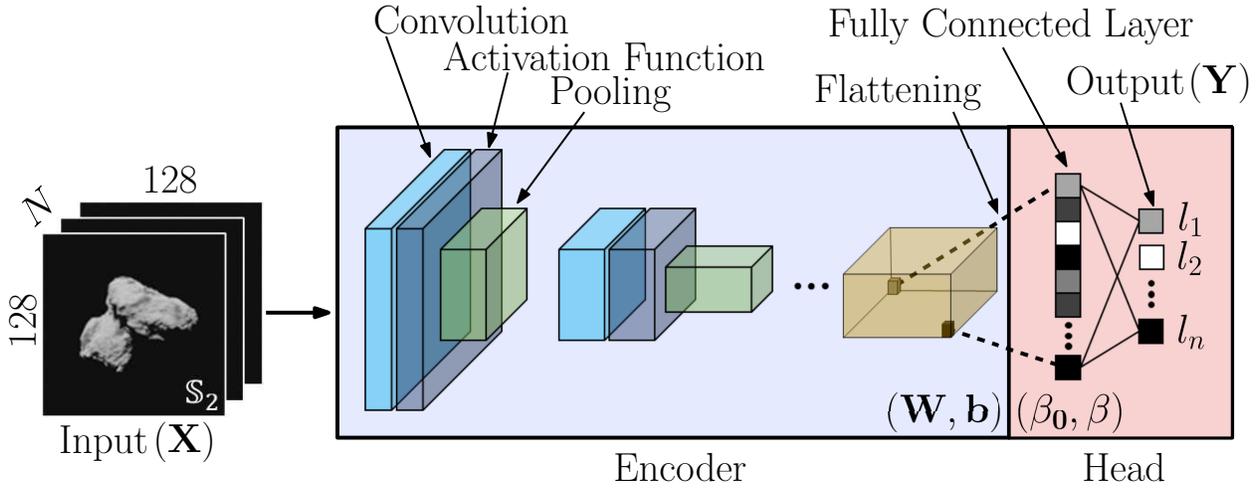}
    \caption{The shared architecture of the \ac{ip} methods used in this work.}
    \label{fig:NN_architectures}
\end{figure}

In this section, the \ac{ip} methods are described in detail with a standardized notation, illustrated in the architecture outlined in \autoref{fig:NN_architectures}. The input $\bm{X}$ is represented by a $128\times128\times N$ tensor while the output $\bm{Y}$ is a vector or matrix whose elements represent the specific labels associated with each image or tensor. 

To simplify the discussion, the architecture is conceptually divided into two main portions: the encoder and the head. In the first one, a hierarchical sequence of convolutions, activation functions, and pooling operations are performed to extract spatial information which is synthesized in the last layer in a latent tensor represented in orange in \autoref{fig:NN_architectures}. Such 3D tensor is flattened into a 2D vector referred to as the fully connected layer which constitutes the first layer of the head of the architecture. In this portion, several hidden layers may or may not be introduced before the output layer. A network is therefore established to map the connection between the neurons from the fully connected layer, which embeds spatial information extracted from the image in the latent space, and the output layer, which expresses the desired output label of the architecture. 

In this work, convolutional layers are driven by weights and biases that only influence the kernels used in the convolution operations. On the other hand, the weights and biases of the head are representing the influence of the neuron connections between the fully connected and output layer. In this work, it is important to consider these two sets of weight and biases separately, since depending on the training strategy, they are handled differently. The former will be referred to as $\bm{W}$ and $\bm{b}$, the latter to $\bm{\beta}$ and $\bm{\beta_0}$. The set of weights and biases of the entire architecture defines the set of parameters referred to as $\bm{\theta} = \left(\bm{W,\;b,\;\beta,\;\beta_0}\right)$. Another parameter used to describe the architecture is represented by $\bm{\Theta}$ which reflects the set of architectural and training choices often referred to as hyper-parameters. 

%These are usually the results of arbitrary choices about the architecture or output of extensive search grid algorithm to finely tune them. 

Within this context, to find an \ac{ip} method which performs well means to find the best desirable sets of $\bm{\Theta}$ and $\bm{\theta}$ which optimize predefined metrics. Training an architecture can be reduced to a parameter estimation problem with a double nested optimization loop. During training, both architectural and training choices (global parameters) and inner weights and biases (local parameters) need to be found out through data. When these are found, they are frozen and at inference the architecture can be considered as a function $\pi$ (parameterized by $\bm{\Theta}$ and $\bm{\theta}$) on the input tensor $\bm{X}$ that generate an output vector $\bm{Y}$: 

\begin{equation}
\bm{Y} = \pi_{\Theta}\left(\bm{X}|\bm{\theta}\right)
\end{equation}

Now that a shared nomenclature is established, the characteristics of the four \ac{ip} methods are described in details.

\subsubsection{Convolutional Neural Network}

In a \ac{cnn} the architecture in \autoref{fig:NN_architectures} is treated as a neural network whose local parameters $\bm{\theta}$ are found using \ac{gd} optimization algorithms. With \ac{gd} all parameters $\bm{\theta}$ are let to vary during training. Several strategies exist depending on the number of images $N_i$ considered in order to perform an update of $\bm{\theta}$. If $N_i=1$ the optimization is referred to as \ac{sgd}; if $N_i=N$ ($N=7500$ in this work) as \ac{gd}; if $1<N_i<N$ it is referred to as \ac{mbgd}. The use of batches makes it possible to load and process smaller tensors, decreasing the computational load and memory while increasing convergence speed, both of which are issues of the \ac{sgd} and \ac{gd}, respectively. The batch size is thus often an important design choice when training a \ac{cnn}.

All the \ac{gd}-based methods are structured in a two-phase process. First, a forward pass is executed by a specific network defined by $\bm{\theta}$ to generate an output. The comparison between such output and its ground-truth value is quantified by a loss metric, which in turn is fundamental in evaluating the performance of the network as well as to determine the necessary adjustments to $\bm{\theta}$ for improvement. These are determined from the computation of the gradient, that is estimated in a backward pass in the network followed by an update of $\bm{\theta}$ to be used in the next iteration. These two passages, especially the last, are computationally expensive, making the training a laborious procedure. A possible way to reduce the training time is to avoid the backward pass, that also require a different optimization scheme.

\subsubsection{Convolutional Extreme Learning Machine}
A \ac{celm} removes the need for a backward pass by solving the optimization problem on the connections between the last hidden layer and the output layer of the architecture.

In this work, the architecture and initialization of \ac{celm} are the same as the one of a \ac{cnn}. However, greater importance is given in the \ac{celm} on the random distribution of the $W $ eightscnn and $\bm{b}$ of the kernels and of $\bm{\beta_0}$. Once they are randomly initialized when the architecture is generated, they are frozen and are not changed during training. This is not the same for $W $ eightscelm, that are the only ones changed during training. Given a set of true input-output samples $\left(\bm{X},\bm{T}\right)$, the forward pass of the input into the network generates a hidden layer output matrix $H$ right before the output layer: 

\begin{equation}
    \bm{H} =\begin{bmatrix} \bm{h}(\bm{x}_1)\\ \vdots \\\bm{h}(\bm{x}_N)\\\end{bmatrix} = 
    \begin{bmatrix} 
    h_1(\bm{x}_1) & \hdots &  h_L(\bm{x}_1) \\
    \vdots & \ddots & \vdots \\
    h_1(\bm{x}_N) & \hdots &  h_L(\bm{x}_N) \\
    \end{bmatrix}
\end{equation}
where $L$ is the dimension of the hidden layer before the output one, represented by the number of neurons in the layer. The training data target matrix is then defined as: 
\begin{equation}
    \bm{T} =\begin{bmatrix} \bm{t}_1\\ \vdots \\\bm{t}_N\\\end{bmatrix} = 
    \begin{bmatrix} 
    t_{11} & \hdots &  t_{1m} \\
    \vdots & \ddots & \vdots \\
    t_{N1} & \hdots &  t_{Nm} \\
    \end{bmatrix}
\end{equation}

To find the best set of weights $\bm{\beta}$ that matches the matrix $\bm{T}$, the following optimization problem shall be solved \cite{Huang2015_CELM}:

\begin{equation}\label{eq:celm_prob}
    \text{Minimize}:       \left\| \bm{\beta} \right\|^{2}_{2} + C\left\| \bm{H\beta}-\bm{T} \right\|^{2}_{2}
\end{equation}

that is a regularized least square that depends on the $C$ $\;$coefficient. The inclusion of first term regarding the minimization of the weights vector $\bm{\beta}$ increases the stability and improves generalization capabilities \cite{Huang2014}. The minimization problem can be solved in an efficient way as: 

\begin{equation}\label{eq:celm_ls}
    \beta = \left\{\begin{matrix}
 \bm{H}^T\left(\frac{\bm{I}}{C} + \bm{HH}^T\right)^{-1}\bm{T}, \quad \text{if} \; N\leq L \\
 \left(\frac{\bm{I}}{C} + \bm{H}^T\bm{H}\right)^{-1}\bm{H}^T\bm{T}, \quad \text{if} \; N > L 
\end{matrix}\right.
\end{equation}
Since the remaining weights and biases ($\bm{W}$, $\bm{b}$, and $\bm{\beta}_0$), are randomly fixed at initialization and are never changed, there is no need for a backward pass. Only the forward pass and the \ac{ls} problem need to be processed, which makes the training of a \ac{celm} architecture much faster than the one of a \ac{cnn}. Note that in this work $\mathbf{\beta}_0$ is not considered in the \ac{celm} architecture. 

\subsubsection{Hybrid Convolutional Extreme Learning Machine}

By combining the design of the \ac{cnn} and the one of the \ac{celm}, the capability of a hybrid \ac{ip} method is also investigated. 

\begin{figure}[hbt!]
    \centering
    \includegraphics[width=1\textwidth]{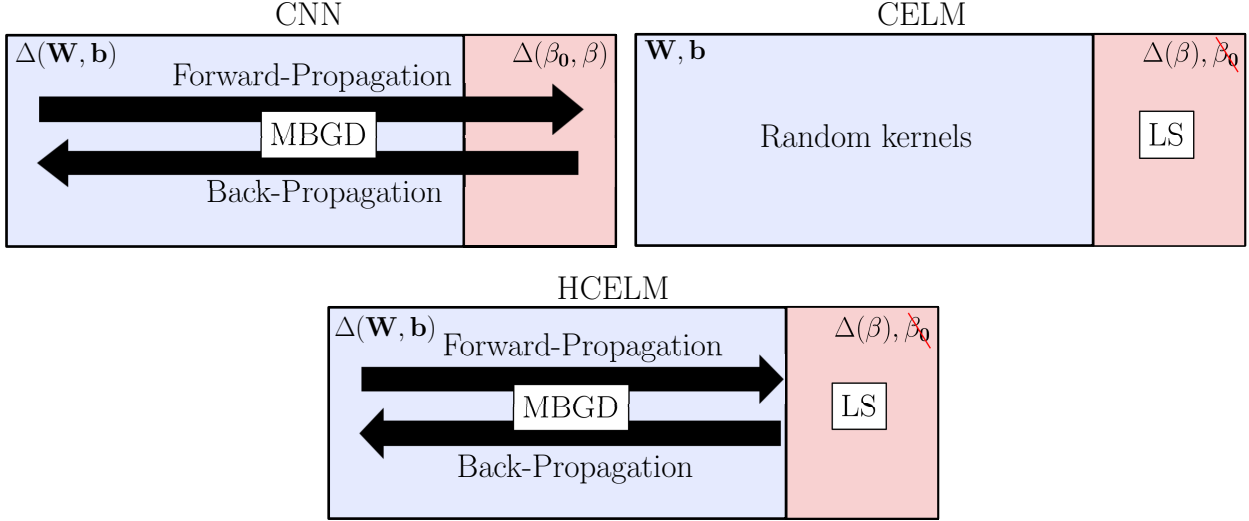}
    \caption{Schematics of the main differences between \ac{celm}, \ac{cnn}, and \ac{hcelm}.}
    \label{fig:Training_NN_architectures}
\end{figure}

In the \ac{hcelm}, transfer learning is used to get the weights and biases of the kernels from the encoder of a previously trained \ac{cnn} architecture, while the head is trained using the \ac{ls} method, as in the \ac{celm} paradigm. Two \ac{ip} methods have been designed with this strategy: one that uses the encoder from the best \ac{cnn} architecture for that dataset (referred to \ac{hcelm}), and one that uses the encoder from the best \ac{cnn} architecture for that reference frame (referred to \ac{hcelm3}). In \autoref{fig:Training_NN_architectures} the schematic difference between the three architectures used in this work is illustrated. 

\subsubsection{Training}

\begin{figure}[htbp]
    \centering
    \includegraphics[width=0.99\textwidth]{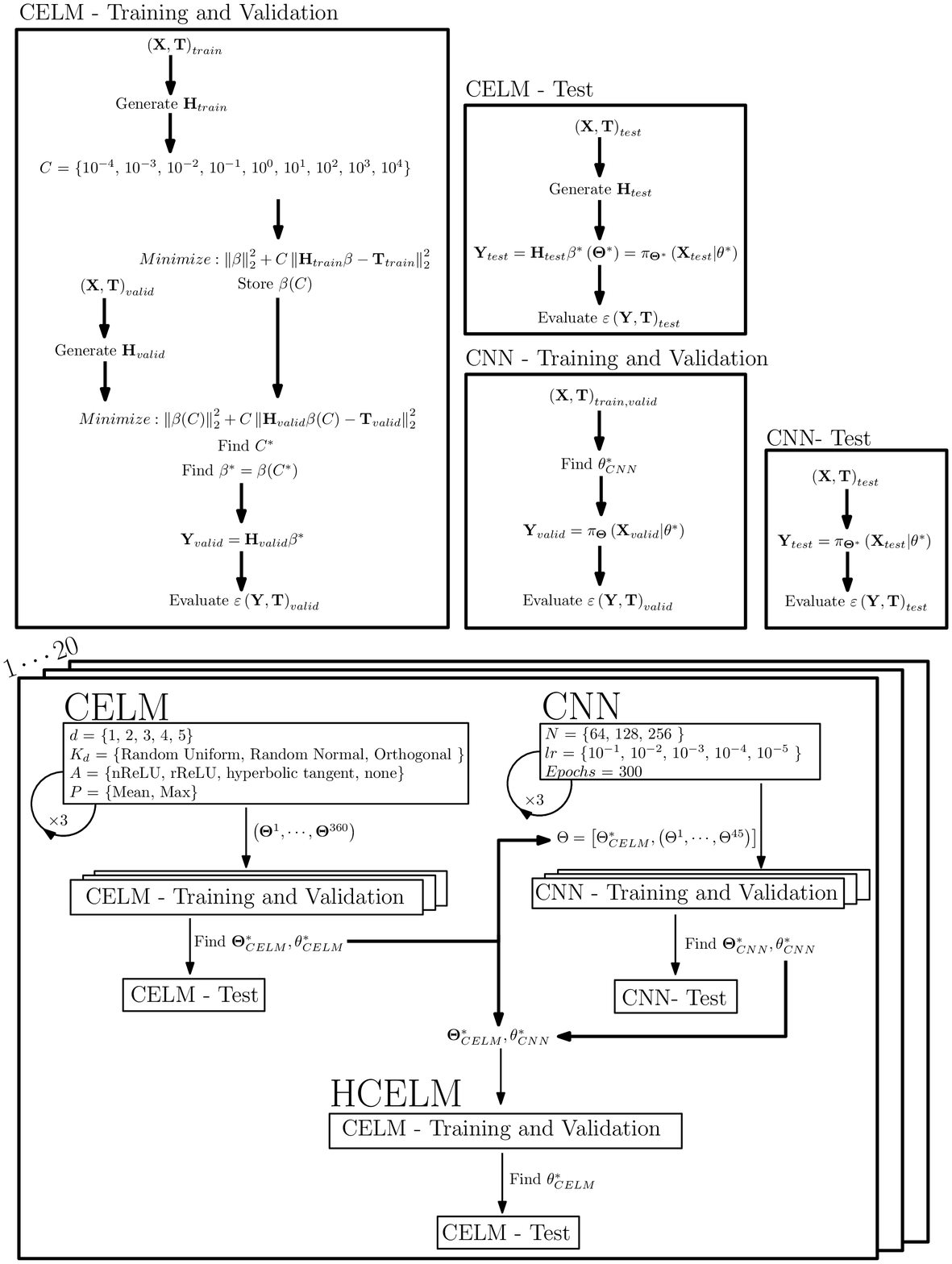}
    \caption{Schematic of the overall training, validation, and testing strategy adopted for all \ac{ip} methods.}
    \label{fig:Training_strategy}
\end{figure}

In this section, the strategy to train the \ac{ip} methods is described. The architecture of each method is generated following a rigid procedural methodology to define the parameters of $\Theta$. A schematic of the procedure is illustrated in \autoref{fig:Training_strategy}.

To begin with, a great number of \ac{celm} architectures are initialized following the same rules for each dataset. The encoder of each architecture is designed with a hierarchical structure \cite{DBLP:conf/icml/SaxeKCBSN11}. While going deeper from input to the fully connected layer, the starting $128\times128\times1$ tensor is squeezed; its size $s_i$ is halved as function of the depth level $i$ as $s_{i} = 2^{7-i}$ while its depth is doubled starting from an arbitrary value of $16$.  

Each depth level $i$ is made up of the consecutive application of a convolution, activation function, and pooling operation. The convolutions are performed with $3\times3$ kernels while the number of kernels $n_{i}^{ker}$ used at each depth is set to increase exponentially as: 

\begin{equation}
    n_{i}^{ker} = 2^{3+i} \qquad n_{0}^{ker} = 1
\end{equation}

The number of weights and biases at each depth level of the encoder can thus be determined: 

\begin{equation}
    n_i^{W} = 9n_{i}^{ker}n_{i-1}^{ker}
    \quad, \quad 
    n_i^{b} = n_{i}^{ker}
\end{equation}

From a network as deep as 1 level to one as deep as 5, the cumulative number of parameters defining the weights and biases of the kernels in the encoder is respectively $160$, $4800$, $23296$, $97152$, and $392320$. Similarly, also the number of weights and biases of the head can be determined. Since the number of neurons in the fully connected layer depends on the depth level as $n_{i}^{fc} = s_{i}^{2} n_{i}^{ker}$, the number of $\bm{\beta}$ and $\bm{\beta}_0$ of the head is computed as:

\begin{equation}
    n_{i}^{\beta} = n_{i}^{fc}n_o \quad, \quad     
    n_{i}^{\beta_0} = n_o
\end{equation}
where $n_0$ is the number of neurons composing the output. For improved handling of the labels, whenever the ($\phi_1,\phi_2,\rho$) labels are considered the azimuth angle $\phi_1$ is transformed in the adimensional $\left( sin\phi_1, cos\phi_1 \right)$ pair. $n_0$ is thus equal to $3$ in all datasets but those for which spherical coordinates are used as labels, for which $n_0=4$. An example of a $5$ layers architecture is illustrated in \autoref{tab:CNN_5} with a parameter count divided per depth.

\begin{table}[htbp]
    \fontsize{10}{8}\selectfont
    \caption{Example of a 5 layers architecture with a 3 output layer.
    }
   \label{tab:CNN_5}
        \centering 
   \begin{tabular}{r  r  l  l} % Column formatting, 
      \hline
      \hline
      \textbf{Layer name} & \textbf{Layer type} & \textbf{Output Shape} & \textbf{Param} \boldmath{$\#$} \\
      \hline
      I & InputLayer & (None, 128, 128, 1) & 0 \\
      \hline
      C1  &  Conv2D  &  (None, 128, 128, 16) & 160 \\
      A1  &  Activation &  (None, 128, 128, 16) & 0 \\
      P1  &  Pooling &      (None, 64, 64, 16) & 0 \\
      \hline
      C2  &  Conv2D  &  (None, 64, 64, 32) & 4640 \\
      A2  &  Activation &  (None, 64, 64, 32) & 0 \\
      P2  &  Pooling &      (None, 32, 32, 32) & 0 \\
      \hline
      C3  &  Conv2D  &  (None, 32, 32, 64) & 18496 \\
      A3  &  Activation &  (None, 32, 32, 64) & 0 \\
      P3  &  Pooling &      (None, 16, 16, 64) & 0 \\
      \hline
      C4  &  Conv2D  &  (None, 16, 16, 128) & 73856 \\
      A4  &  Activation &  (None, 16, 16, 128) & 0 \\
      P4  &  Pooling &      (None, 8, 8, 128) & 0 \\
      \hline
      C5  &  Conv2D  &  (None, 8, 8, 256) & 295168 \\
      A5  &  Activation &  (None, 8, 8, 256) & 0 \\
      P5  &  Pooling &      (None, 4, 4, 256) & 0 \\
      \hline
      FC  &  Flattening  &  (None, 4096) & 0 \\
      O   &  Dense  &  (None, 3) & 12291 \\
      \hline
      \hline
   \end{tabular}
\end{table}

%160, 4640, 18496, 73856, 295168
%160, 4800, 23296, 97152, 392320
Having defined a procedural set of rules to generate each architecture, these are generated with the hyper-parameters $\Theta$ summarized in \autoref{tab:NN_parameters}.

\begin{table}[htbp]
\caption{Sets of $\bm{\Theta}$ explored for \ac{celm} and \ac{cnn} methods.}
\label{tab:NN_parameters}
\centering
\begin{tabular}{p{3cm} p{1cm} p{5cm} p{4cm}}
\hline
\hline
\textbf{Parameter} & \textbf{Symbol} & \textbf{Description} & \textbf{Possible values} \\ 
\hline
Number of layers & $d$ & Number of hidden layers in the architecture  & 1, 2, 3, 4, 5 \\

Kernel distribution & $K_d$ & Random distribution of the weight and biases of the kernels & Random Uniform (-1,1),  Random Normal (0, 1), Orthogonal \\

Activation function & $A$ & Activation function used after the convolution operation & \ac{nrelu}, \ac{relu}, \ac{tanh}, none\\

Pooling strategy & $P$ & Pooling strategy after the activation function & Mean, Max\\

Regularization coefficient & $C$ & Regularization coefficient of \autoref{eq:celm_prob} & $10^{-4}$, $10^{-3}$, $10^{-2}$, $10^{-1}$, $10^{0}$, $10^{1}$, $10^{2}$, $10^{3}$, $10^{4}$ \\
\hline
Batch size & $N$ & Batch size used in the \ac{mbgd} & 64, 128, 256 \\
Learning rate & $lr$ & Learning rate used in the \ac{mbgd} & $10^{-1}$, $10^{-2}$, $10^{-3}$, $10^{-4}$, $10^{-5}$\\
\hline
\hline
\end{tabular}
\end{table}

By considering the possible combination between $d$, $K_d$, $A$, and $P$ a total of 120 different \ac{celm} architectures are considered. For each kernel distribution, a random initialization is executed 3 times, meaning that a total of 360 \ac{celm} networks are generated for each dataset in \autoref{tab:notation_all}.

Once the forward pass of the \ac{celm} is executed, for each architecture the \ac{ls} optimization problem is run 9 different times during training of the \ac{celm} with different regularization terms $C$. The training set is used to determine all possible values of $\bm{\beta}$ depending on $C$$ $, while the validation set is used to determine the best value of $C$. The combination of $\bm{\beta}$ determined from the training set and $C$$ $ determined from the validation set is used in inference on the test set to produce the estimated labels. Because the training time of \ac{celm} is orders of magnitude faster than the one of the \ac{cnn} ($\sim 1 s$ compared to $\sim 600 s$, on average), with the use of \ac{celm} networks it is possible to explore the architecture space to find those that seems inherently more suitable for the task at hand. As suggested in \cite{DBLP:conf/icml/SaxeKCBSN11}, a great portion of the performance of the \ac{cnn} seems to be given by the architectural choices, which are often neglected nor sufficiently explored given the large training time required by \ac{gd} methods. 

For each dataset, the best \ac{celm} network is defined as the one achieving the minimum positioning error $\varepsilon_n$ (defined in \autoref{sec:results}) on the test set while its parameters $\bm{\Theta}$ are saved.

For each dataset, a \ac{cnn} architecture is then initialized with the hyper-parameters $\bm{\Theta}$ found from the \ac{celm} architectures. This \ac{cnn} is then trained by varying $N$$ $ and $lr$$ $ and performing 3 runs for each combination, for a total of 45 cases. For each of them, the \ac{cnn} is trained for $300$ epochs while the best value on the validation loss is used to instantiate the weights and biases of the best possible realization of the \ac{cnn}. 

In this way, we provide the \ac{cnn} with an architecture that has proven to work properly even with random kernels and investigate whether the re-arrangement of the weights and biases of these kernels together with the ones in the head portion can further improve the baseline performances by extracting additional information or not. The set of $\bm{\Theta}$ for the best \ac{celm} and \ac{cnn} architectures is summarized in \autoref{tab:CELM_CNN_20}.

%The total training time for the 45 different cases of \ac{cnn} takes roughly 10 minutes for each \ac{cnn}, roughly 7.5 hours for each dataset. 

\begin{table}[htbp]
\caption{Best sets of $\bm{\Theta}$ found during training and used in inference.}
\label{tab:CELM_CNN_20}
\centering
\begin{tabular}{l | ccccc | cc}
\hline
\hline
\multirow{2}{*}{\textbf{Dataset}} & \multicolumn{5}{c|}{\textbf{CELM parameters}} & \multicolumn{2}{c}{\textbf{CNN parameters}} \\
 & \textbf{$d$} & \textbf{$K_d$} & \textbf{$A$} & \textbf{$P$} & \textbf{$C$} & \textbf{$N$} & \textbf{$lr$} \\
\hline
 D1 & 5 & RandomUniform & tanh & Mean & $10^{-1}$ & 64 & $10^{-3}$\\
 H1 & 5 & Orthogonal & relu & Mean & $10^{1}$ & 256 & $10^{-4}$\\
 L1 & 5 & RandomNormal & tanh & Mean & $10^{-2}$ & 64 & $10^{-3}$\\
 P1 & 5 & RandomNormal & tanh & Mean & $10^{-2}$ & 64 & $10^{-3}$\\
 \hline
 D2 & 5 & Orthogonal & nrelu & Mean & $10^{1}$ & 64 & $10^{-3}$\\
 H2 & 5 & Orthogonal & relu & Mean & $10^{1}$ & 64 & $10^{-3}$\\
 L2 & 5 & Orthogonal & relu & Mean & $10^{1}$ & 64 & $10^{-3}$\\
 P2 & 5 & Orthogonal & nrelu & Mean & $10^{1}$ & 64 & $10^{-3}$\\
 \hline
 D3 & 5 & Orthogonal & none & Max & $10^{-2}$ & 64 & $10^{-3}$\\
 H3 & 5 & Orthogonal & nrelu & Mean & $10^{1}$ & 64 & $10^{-3}$\\
 L3 & 5 & Orthogonal & relu & Mean & $10^{1}$ & 64 & $10^{-3}$\\
 P3 & 5 & Orthogonal & nrelu & Mean & $10^{1}$ & 64 & $10^{-3}$\\
 \hline
 D4 & 5 & Orthogonal & relu & Mean & $10^{2}$ & 64 & $10^{-3}$\\
 H4 & 4 & Orthogonal & relu & Mean & $10^{1}$ & 64 & $10^{-3}$\\
 L4 & 5 & Orthogonal & nrelu & Mean & $10^{1}$ & 64 & $10^{-3}$\\
 P4 & 5 & Orthogonal & nrelu & Mean & $10^{1}$ & 64 & $10^{-3}$\\
 \hline
 D5 & 5 & Orthogonal & nrelu & Mean & $10^{1}$ & 64 & $10^{-3}$\\
 H5 & 5 & Orthogonal & relu & Mean & $10^{1}$ & 64 & $10^{-3}$\\
 L5 & 5 & Orthogonal & relu & Mean & $10^{1}$ & 64 & $10^{-3}$\\
 P5 & 5 & Orthogonal & relu & Mean & $10^{1}$ & 64 & $10^{-3}$\\
\hline
\hline
\end{tabular}
\end{table}

The other two possible setups are then further investigated with the \ac{hcelm} architecture. The final \ac{cnn} encoders are frozen into their architectures, which are re-trained as \ac{celm} by changing only the $\bm{\beta}$ in the head portions. The training is the same for the \ac{celm} ones, but performed only on one architecture and not on $360$ ones. First, the encoder for each \ac{cnn} is considered, then only the best encoders are shared amongst all networks which work in the same reference frame. This means that in the first case 20 encoders are used, one for each architecture, then 3 encoders are used, each of them being the representative of the \ac{cnn} architecture working best in that reference frame. 

All \ac{ip} methods are considered to work with normalized input and output, which has been observed to improve the overall performance. The labels are normalized over the maximum and minimum values found in each training set. While \ac{celm} are trained with the entire dataset at once, \ac{cnn} architectures are trained with batches of images. In the case of \ac{cnn} architectures, this was due to hardware limitations, while the number of $7500$ images for the training set similarly comes from hardware limitations related to matrix inversion in the \ac{celm} training. For the \ac{cnn} one epoch is considered when all batches are processed. 300 epochs are considered for the \ac{cnn} while by definition the training of the \ac{celm} happens in one epoch. Adam is used as optimization algorithm to train the methods requiring \ac{mbgd}.

\section{Results}\label{sec:results}

The four \ac{ip} methods trained over all datasets of \autoref{tab:notation_all}, namely \ac{celm}, \ac{cnn}, \ac{hcelm}, and \ac{hcelm3}, represented by the hyper-parameters $\Theta$ detailed in \autoref{tab:CELM_CNN_20} are applied in inference over the $5000$ images of the corresponding test sets. For simplicity, a consistent colormap is used to distinguish the four methods. Purple, blue, green, and yellow are used respectively for the \ac{celm}, \ac{cnn}, \ac{hcelm}, and \ac{hcelm3} architectures. In order to effectively synthesize the performances of all the techniques considered, the following navigation error metrics are defined:

\begin{equation}
    \bm{\varepsilon}_p^i = \bm{p}^{i, \mathbb{S}_0}_{est}-\bm{p}^{i, \mathbb{S}_0}_{true}
\end{equation}

\begin{equation}
    \varepsilon_n = \frac{
    \left\|
    \bm{\varepsilon}_p^i
    \right\|_2
    }{\rho_{true}^{\mathbb{S}_0}}100
\end{equation}
where $\bm{p}^{i, \mathbb{S}_0}_{est}$, $\bm{p}^{i, \mathbb{S}_0}_{true}$ are respectively the estimated and true position in the $i$ reference frame while $\rho_{true}^{\mathbb{S}_0}$ is the true range from the body \ac{com}, all evaluated in $\mathbb{S}_0$. Also, for the architectures that work with the $\left(\bm{\delta}, \rho \right)$ labels, the following additional metrics are defined: 
\begin{equation}
    \varepsilon_{CoF}^{u} = CoM^{u}-CoF^{u} \quad,\quad  \varepsilon_{CoF}^{v} = CoM^{v}-CoF^{v} 
\end{equation}

\begin{equation}
    \varepsilon_{CoF} = \sqrt{\left(\varepsilon_{CoF}^{u}\right)^2+\left(\varepsilon_{CoF}^{v}\right)^2}
\end{equation}

\begin{equation}
    \varepsilon_{\rho} = \rho_{est}-\rho_{true}
\end{equation}

\begin{figure}[hbt!]
    \centering
    \includegraphics[width=1.35\textwidth, angle = -90]{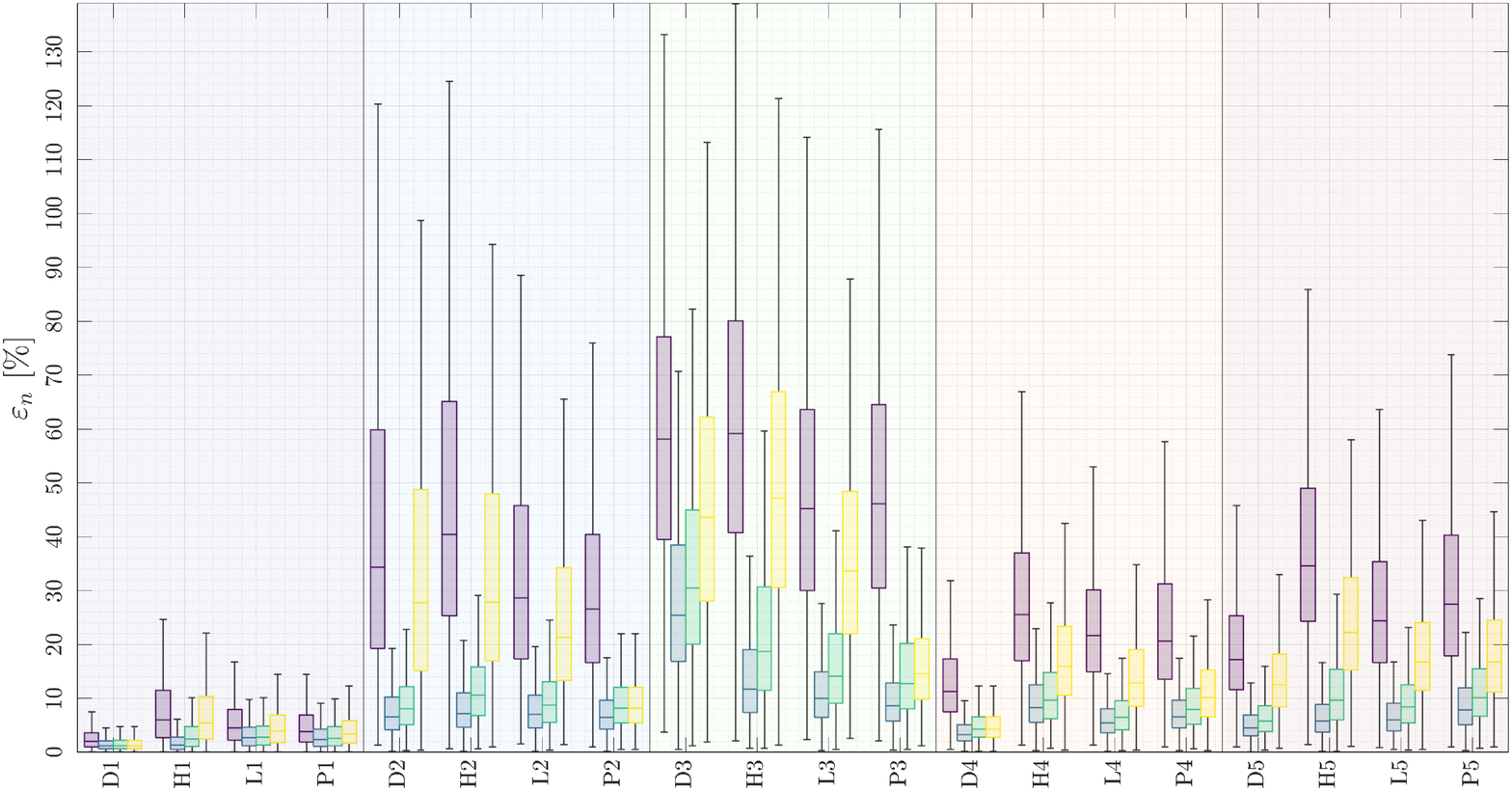}
    \caption{Box plot of $\varepsilon_n$ for all cases considered.}
    \label{fig:Boxplot}
\end{figure}

\begin{figure}[hbt!]
    \centering
    \includegraphics[width=1\textwidth]{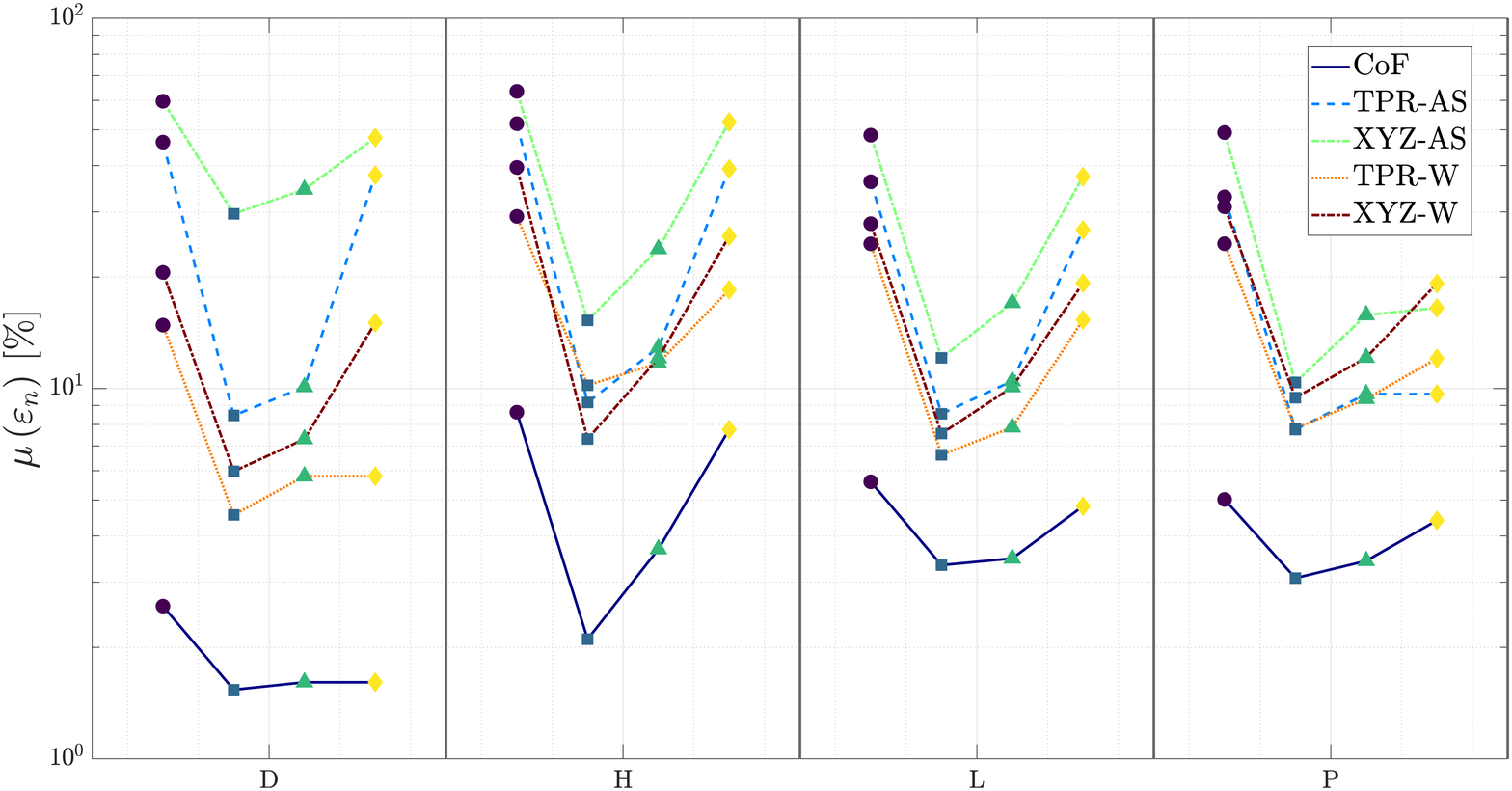}
    \caption{Summary plot of the mean $\varepsilon_n$ values achieved by all possible combinations of the dataset and architecture investigated in this work.}
    \label{fig:semilogy_performance}
\end{figure}

\begin{figure}[hbt!]
    \centering
    \includegraphics[width=1\textwidth]{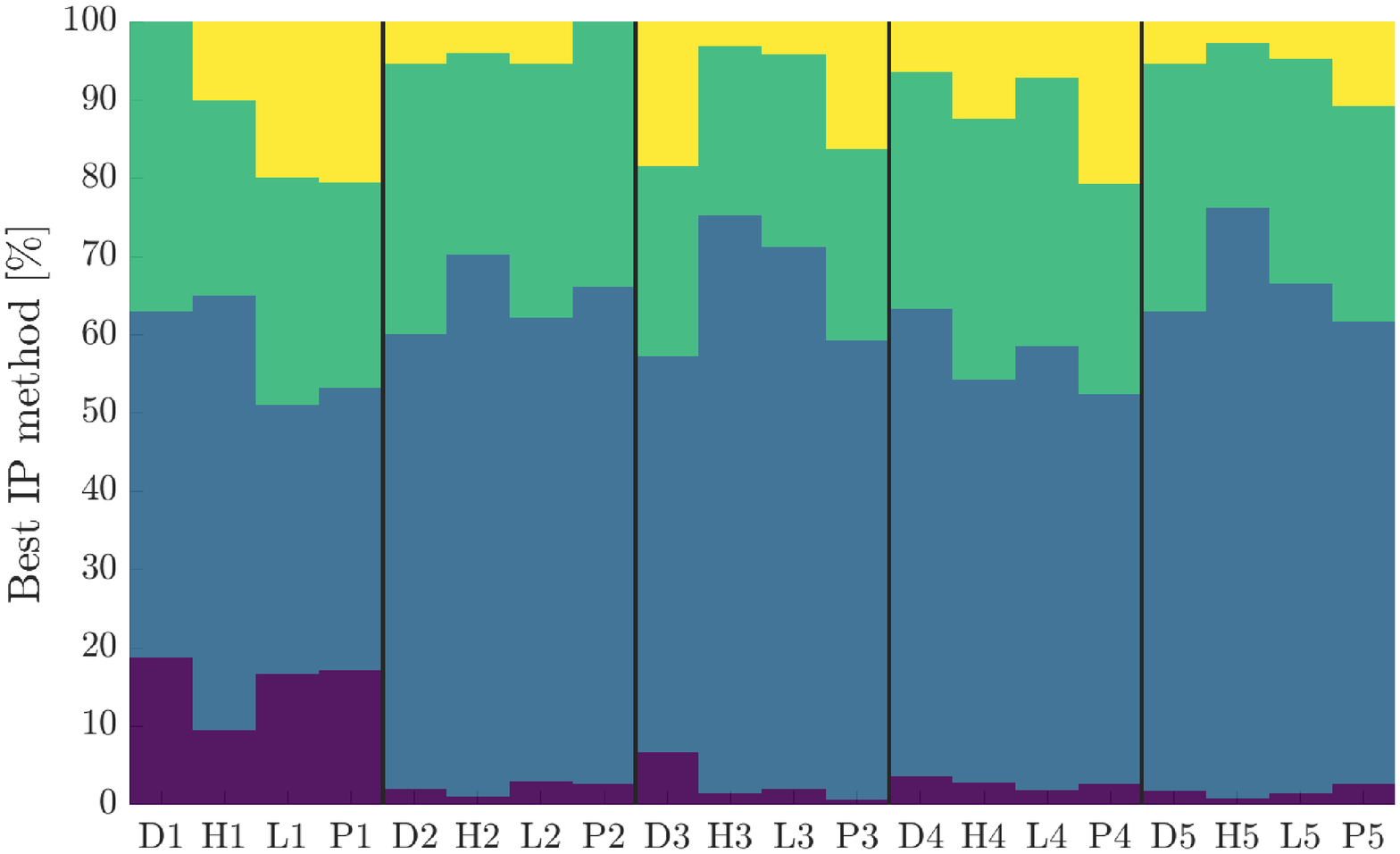}
    \caption{Shares of the best \ac{ip} method across different datasets.}
    \label{fig:best_plot}
\end{figure}

From \autoref{fig:Boxplot}, \autoref{fig:semilogy_performance}, and \autoref{fig:best_plot} global performance in terms of $\varepsilon_n$ are summarized for all cases considered. From these plots, it is possible to draw important considerations on the importance of labeling strategy, reference frame, and training method.

In \autoref{fig:Boxplot} the box plot are organized from top-down in groups of 5 by macro-categories based on the labeling strategy and then in clusters of 4 based on the \ac{ip} method, ordered from top-down as \ac{celm}, \ac{cnn}, \ac{hcelm}, and \ac{hcelm3}. It is possible to immediately appreciate that all the methods trained with the labeling strategy based on ($\bm{\delta}, \rho$) outperform all others considered. This is especially remarkable for the \ac{celm} method, which is performing similarly to the other \ac{ip} methods when considering this labeling strategy. On the other hand, in all other cases, \ac{celm}-based methods perform poorly. This is especially true when compared with the \ac{cnn} ones, which outperform all methods considered in all possible combinations. Focusing on the importance of the reference frame, it is observed that overall better performances are achieved when the labels are expressed in the $W $  reference frame than in the $AS $  one, while the choice of the coordinate system (cartesian or polar) seems to cause only minor differences. 

An even more concise representation is visible in the semi-log plot of the mean $\varepsilon_n$ error clustered by body in \autoref{fig:semilogy_performance}. Once again it is possible to appreciate the much better performance achieved with the ($\bm{\delta}, \rho$) labeling strategy compared to the other ones for all bodies considered. In such a case, the performance of the \ac{celm} method is not so different than the one of the \ac{cnn}. While the former have a mean $\varepsilon_n$ of 2.58, 8.63, 5.60, 5.02 (respectively for \ac{d}, \ac{d}, \ac{h}, and \ac{p}), the latter generates position estimates only 1.68, 4.11, 1.68, and 1.63 times better. Apart from \ac{h}, this means that only a very marginal performance gain is achieved with the use of a \ac{cnn} rather than a \ac{celm}. This does not hold when comparing \ac{celm} and \ac{cnn} performances for other labeling strategies, which show much wider gaps. From \autoref{fig:semilogy_performance} it is also possible to observe a trend depending on the shape considered: simpler, regular shapes such as the one of Didymos (\ac{d}) are better exploited for navigation than highly irregular ones such as Lutetia (\ac{l}), 67P (\ac{p}), and Hartley (\ac{h}). 

Finally, \autoref{fig:best_plot} reports in a stacked histogram plot the share for each dataset in which an \ac{ip} method performs better than all others considered. While the \ac{cnn} is always the best candidate, \ac{hcelm} consistently scores as the second best method across all datasets considered. Moreover, the \ac{celm} is considered the best third option only in the case of the ($\bm{\delta}, \rho$) labeling strategy. This graphic representation is possible thanks to the fact that the geometric points considered across the test sets are the same.  

%2.58, 8.63, 5.60, 5.02
%1.54, 2.10, 3.33, 3.08

\begin{figure}[hbt!]
    \centering
    \includegraphics[width=1\textwidth]{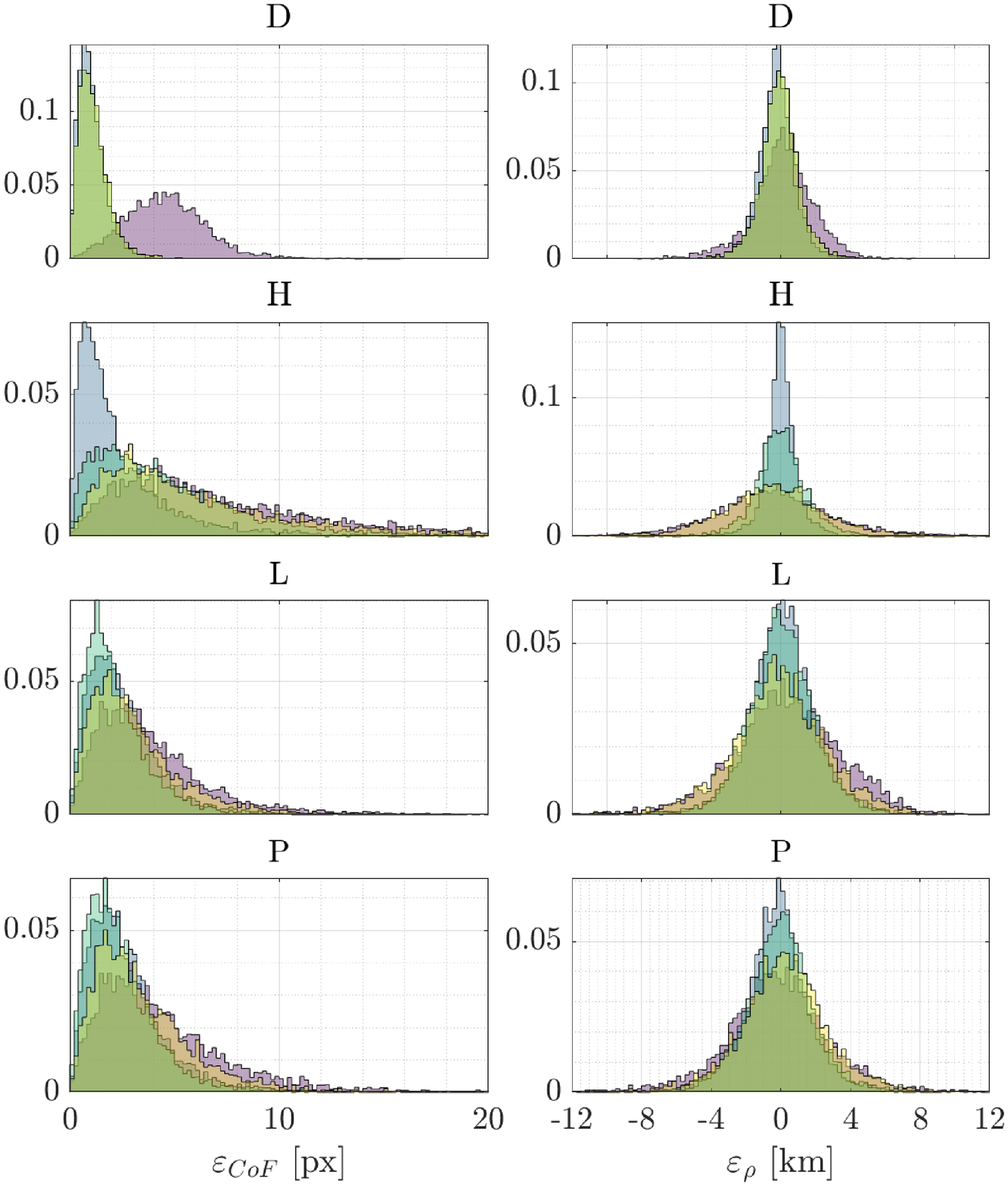}
    \caption{Normalized histograms of the $\varepsilon_{CoF}$ (left) and $\varepsilon_{\rho}$ (right) on different bodies with \ac{ip} methods considered.}
    \label{fig:histogram_S2}
\end{figure}

Since it is observed that the labeling strategy based on ($\bm{\delta}, \rho$) works better than the others, the performance is further investigated in this case. In \autoref{fig:histogram_S2} the histograms of the $\varepsilon_{CoF}$ and $\varepsilon_{\rho}$ metrics in $\mathbb{S}_2$ space are illustrated. When considering \ac{d}, it is commented that the \ac{celm} shows a much larger mean error in $\varepsilon_{CoF}$ than the other methods while only a smaller variability both in terms of means and variance is observed in $\varepsilon_{\rho}$. The fundamental failure mechanism for which the \ac{celm} method performs worse than the \ac{cnn} one with \ac{d} is thus represented by a larger mean error in the estimated \ac{cof} coordinates. When considering \ac{l} and \ac{p}, the difference between the histograms of the various methods is more subtle since only small variations in the mean and variance are observed across the different methods. On the other hand, when considering \ac{h}, it is possible to see that \ac{cnn} is much more accurate both in the \ac{cof} and range estimates than all other methods. The hybrid \ac{celm} performs better in the range estimate (with similar variance than the \ac{cnn}) but does not perform at the same level as the \ac{cnn} in the \ac{cof} estimate.

\begin{figure}[hbt!]
    \centering
    \includegraphics[width=1\textwidth]{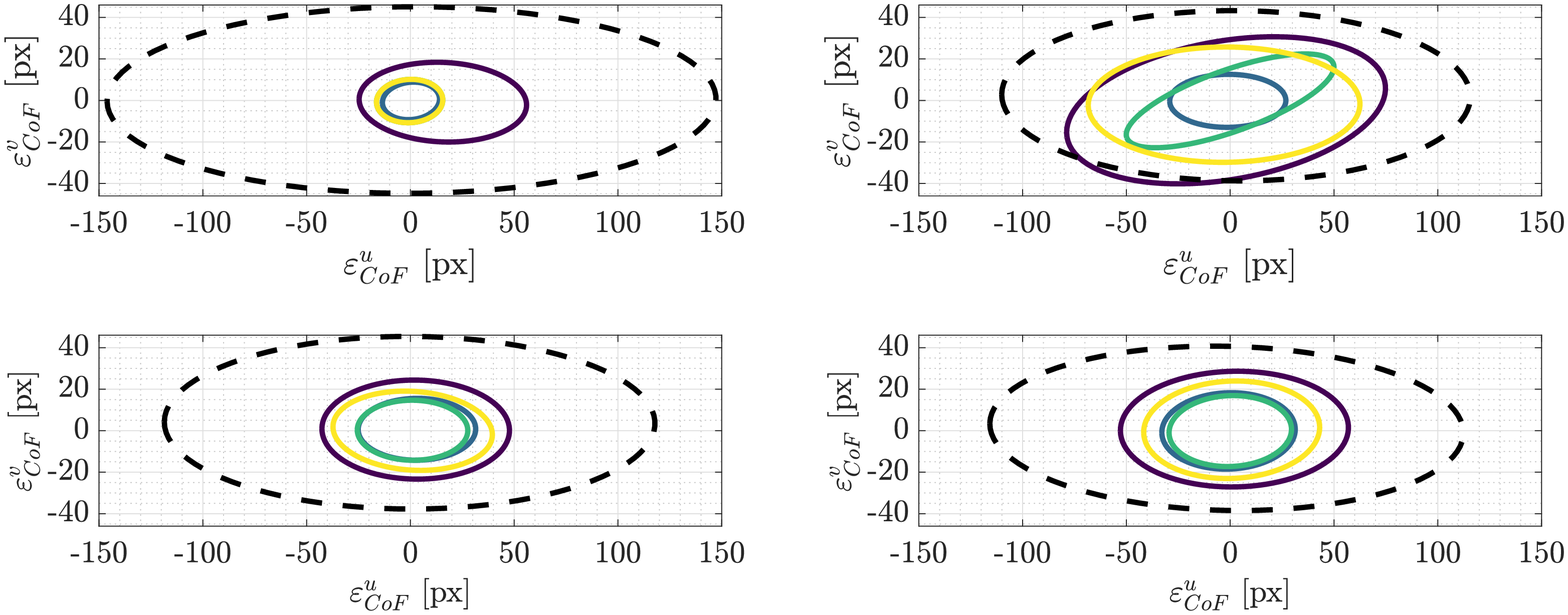}
    \caption{Error ellipses in $UV $  frame and $\mathbb{S}_0$ space on different bodies with \ac{ip} methods considered.}
    \label{fig:ellipses_S0}
\end{figure}

In \autoref{fig:ellipses_S0} the error ellipses of the \ac{cof} coordinate in $\mathbb{S}_2$ space are illustrated together with the error ellipse (dashed dark ellipse) that would have been obtained by not correcting the \ac{cob} with a data-driven scattering law implemented by the \ac{ip}. It is also observed that different than all other ellipses, the \ac{celm} ellipse with \ac{d} is not centered in the zero error point thus introducing a bias that is causing the larger mean error already observed in \autoref{fig:histogram_S2}. 

\begin{figure}[hbt!]
    \centering
    \includegraphics[width=1\textwidth]{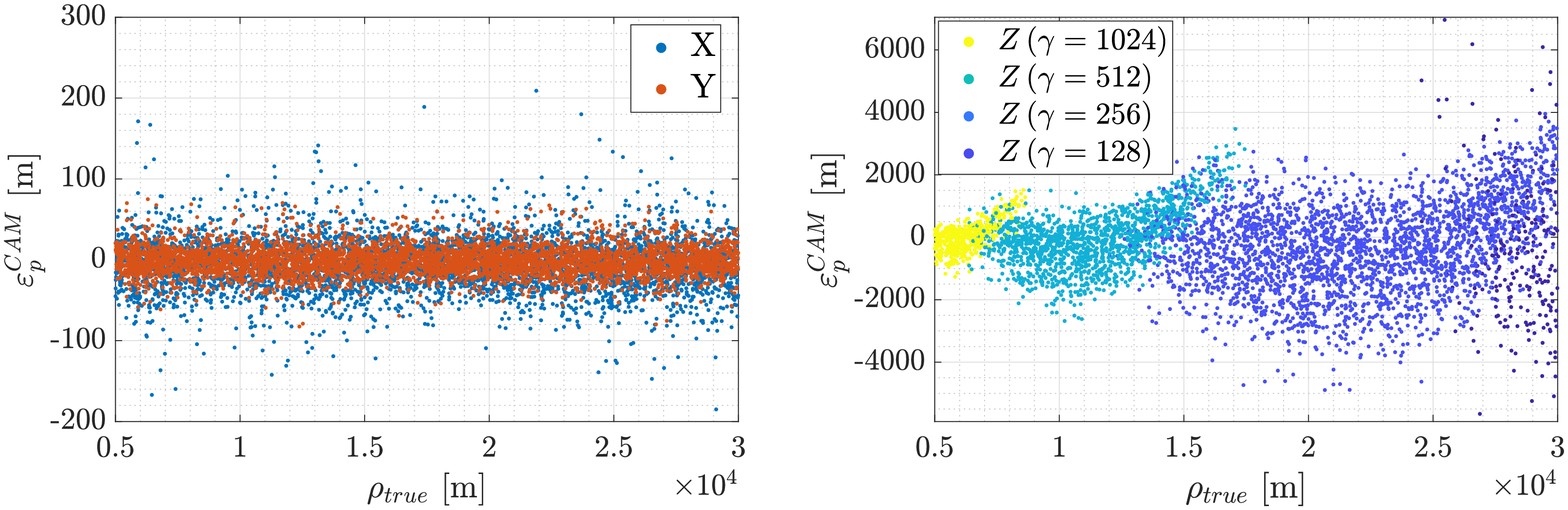}
    \caption{Position error by components in the $CAM $ reference frame for the CELM on \ac{l}.}
    \label{fig:error_CAM}
\end{figure}

In \autoref{fig:error_CAM} a representative case of the position error of the \ac{celm} in $CAM $ reference frame for the \ac{l} body is illustrated. It is possible to see that the position error is one order of magnitude higher in the boresight direction of the camera than on the other axis. This result is expected from optical-based navigation systems. This means that the error in the range estimate from the body is the major contributor to the positioning performances when using this labeling strategy. 

\begin{figure}[hbt!]
    \centering
    \includegraphics[width=1\textwidth]{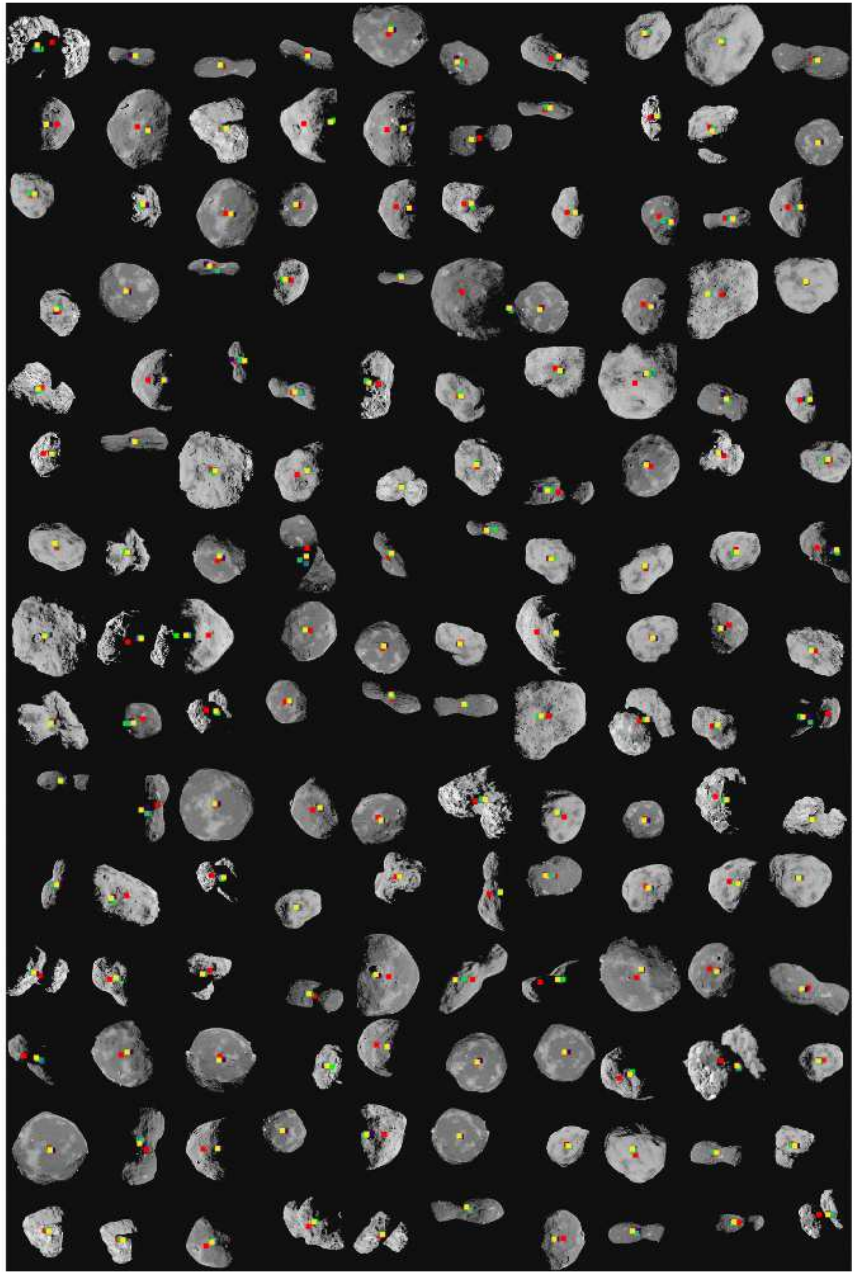}
    \caption{Sample of \ac{cof} estimates on images from the test set with \ac{cob} (red) and \ac{com} (green) visualized.}
    \label{fig:Mosaic}
\end{figure}

Eventually, a visualization of a sample of $150$ images in $\mathbb{S}_2$ and the \ac{cof} prediction by the different \ac{ip} methods is illustrated in \autoref{fig:Mosaic}. 

\clearpage

\section{Conclusions}\label{sec:conclusions}
In this work, \ac{celm}s are investigated as a possible alternative to \ac{cnn}s for autonomous vision-based navigation systems around small bodies. This is done in an extensive analysis considering 4 different small body shapes and 5 different labeling strategies, for a total of 20 scenarios, each of which is examined with 4 different \ac{ip} methods, resulting in tens of thousands of different architectures explored. 

It is demonstrated that the coupling between \ac{ip} method and labeling strategy plays a fundamental role to achieve the desired performance. In particular, the ($\bm{\delta}, \rho$) labeling strategy is found to be the best one. In this case, the navigation labels are given by two geometrical quantities which are estimated directly from the image, and a third, more difficult, label associated with the range from the body. The error on such a label is the one driving the error in the position estimate. 

It is also noted that amongst the other labeling strategies, the ones adopting the $W $  reference frame perform better than those using the $AS $  one. This hints at the fact that in the former the estimated position relies on the correlation with global geometric properties of the body which are not dependent on the specific rotational state but are rather dependent on the illumination conditions. On the contrary, in the latter, the methods need to learn illumination invariant features that correlate to the specific rotational state of the body. This seems to be much harder to do, at least when $7500$ images are considered for training. Future work will be focused on changes in the method to allow larger datasets to be used. Irrespective of the reference frame, it is commented that the mapping from images to polar or cartesian coordinates seems to degrade the performance. In particular, while the ($\bm{\delta}, \rho$) and ($\phi_1$, $\phi_2$, $\rho$) labeling strategies both share $\rho$ as a label, the latter performs considerably worse than the former. This may hint at a more difficult mapping from images into two angles in spherical coordinates than the mapping to projective quantities in the image plane, the latter correlating much easily and directly from image. 

Although the \ac{cnn} outperform all other methods considered, the \ac{celm} seems a credible alternative, since it performs in the same order of magnitude as the \ac{cnn} method when considering the best labeling strategy.

It is also commented that the usage of \ac{celm} for exploration of the architecture space generated \ac{cnn} architectures that experienced robust training. This bootstrap training strategy could be deployed as a standard and efficient way to explore the global architecture space as suggested in \cite{DBLP:conf/icml/SaxeKCBSN11}. Moreover, consistently with the findings in \cite{DBLP:conf/icml/SaxeKCBSN11,Huang2014,Huang2015_ELM,Huang2015_CELM}, a preference is observed for orthogonalization of the weights and biases of the kernels, as it is possible to see from the best hyper-parameters $\bm{\Theta}$ in \autoref{tab:CELM_CNN_20}.

Future works will address the addition of dropout into the networks to generate uncertainty estimates, algorithmic design choices to adapt the \ac{celm} for small body images, larger datasets, and the usage of segmentation maps as input in place of grayscale images, the former representing easier input to work with.

\section*{Acknowledgments}
The authors would like to acknowledge the funding received from the European Union’s Horizon 2020 research and innovation programme under the Marie Skłodowska-Curie grant agreement No 813644. A special thanks to Prof. Roberto Furfaro, whose seminar at the Deep-Space Astrodynamics Research \& Technology Group on the application of \ac{elm} theory in Physics Informed Neural Networks contributed to the main idea behind this paper. 

\FloatBarrier

\bibliography{main}

\end{document}